\documentclass[10pt, conference, compsocconf]{IEEEtran}
\usepackage{float}
\usepackage{graphicx}
\ifCLASSINFOpdf
\else
\fi
\begin{document}
%
\title{Genetic Stereo Matching Algorithm with Fuzzy Fitness}


\author{\IEEEauthorblockN{Haythem Ghazouani}
\IEEEauthorblockA{Department of Computer Science\\
\'Ecole Sup\'erieure de Technologie et d'Informatique\\
Charguia II, Tunis 2035, Tunisia\\
Universit\'e de Carthage\\
haythemghz@yahoo.fr}

}


%


\maketitle

\begin{abstract}
This paper presents a genetic stereo matching algorithm with fuzzy
evaluation function. The proposed algorithm presents a new encoding
scheme in which a chromosome is represented by a disparity matrix.
Evolution is controlled by a fuzzy fitness function able to deal with
noise and uncertain camera measurements, and uses classical evolutionary
operators. The result of the algorithm is accurate dense disparity
maps obtained in a reasonable computational time suitable for real-time
applications as shown in experimental results.

\end{abstract}

\begin{IEEEkeywords}
Dense stereo matching; Fuzzy fitness; Genetic algorithm; Disparity map;

\end{IEEEkeywords}

%
\IEEEpeerreviewmaketitle

\section{Introduction}
Evolutionary algorithms try to solve complex problems by imitating the Darwinian evolution process.
In an EA, artifi{}cial critters are created to search for a solution over the space problem. The artifi{}cial critters, known as individuals are constantly competing with each other to discover optimal areas of the search space. Each individual is defined by a fixed encoding scheme representing a single possible solution to the problem. The EA is started by creating an initial population of size $\mu$ including randomly generated individuals {[}8{]}. An evaluation score, known as fitness, is then assigned to each individual. Fi{}tness is generated so that the individual approaches a potential solution as well as possible.
After this initial step, the evolutionary algorithm goes into the main iterative cycle. It produces 
$\lambda$ children from the $\mu$ individuals in current population. The children are produced using perturbation of individual encoding (mutation) and recombination between two or more individuals (crossover). All individuals (children) newly created are assigned fi{}tness scores. A second generation of population of $\mu$ individuals is formed from $\mu$ individuals in the current population and the $\lambda$ children. The same iterative cycle is applied to the new generation and all the generations that will follow. During the whole cycle, an adaptation pressure is applied to individuals. That is, evolutionary approach of the survival of the fi{}ttest is applied and individuals try to outrace each other. The adaptation pressure is done by selection, with the fi{}tter individuals more likely to be chosen for the next generation. The selection pressure is applied when designating individuals for crossover or when electing individuals to build a new population {[}8{]}.

Genetic algorithms (GAs) are adaptive search heuristics that belong to the family of evolutionary algorithms. As such, they use natural selection and genetics to perform a random search in order to solve optimization problems. Although based on random processes, genetic algorithms are not entirely arbitrary, instead they use fitness scores to  guide the search towards areas of better performance within the search space. The key techniques of GAs are performed to mimic the natural evolutionary processes, particularly those adopt the Darwinian rule of "survival of the fittest". Since in nature, the competition between individuals to achieve results in terms of scarce resources results in the fittest individuals surviving at the expense of the weaker ones {[}16{]}.

The application of artificial genetic algorithms in various fields such as image processing, pattern recognition or machine learning has yielded encouraging results {[}1, 9{]}. In particular, several genetic approaches have been used to solve the problem of stereo matching {[}2, 3, 4{]}. To obtain a disparity map, these methods use fitness functions defined from similarity and disparity smoothness constraints. They start by generating an initial population of individuals (chromosomes). Each individual encodes a possible match with respect to local constraints. After that, the evolutionary process is launched to reach a solution for which the pairing is as compatible as possible with respect to matching constraints. Genetic stereo matching algorithm provides good results, but its major drawback is the computational effort required to achieve a satisfactory solution. This disadvantage can be explained by the fact that classical genetic approaches use binary encodings for individuals, which leads to some matching ambiguities {[}5{]}. Furthermore, a binary encoding requires more storage space and more computing time. To get around these limitations, we propose a new encoding for individuals, more compact than binary encoding and requiring much less space. Reducing the storage space has the immediate effect of reducing the computational time and allows the search algorithm to explore the search space more efficiently {[}5{]}. Thus, the convergence time is considerably
improved. The proposed algorithm uses also a fuzzy fitness, to overcome
noise and uncertain camera measurement problems. In the next section,
we present the proposed algorithm.

\section{The genetic fuzzy stereo-matching algorithm}

In stereo-vision problems, we have a pair of images depicting the same scene, taken from two different angles of view. The goal is to determine the corresponding pixels (the projections of the same punctual area on both images). The first idea that comes to the mind is comparing the areas around the two pixels to have a similarity score. Once the similarity score calculated, the result can be improved by including restrictions and calculating the matching that maximizes the global similarity {[}6, 14, 15{]}.
The epipolar restriction is used to reduce the search space {[}6{]}. 

Genetic algorithms are adaptive search methods based on the principles of evolution theory, i.e. crossover, mating, mutation, natural selection, fitness. These principles are adapted to be applied in the context of stereo matching.

Before defining genetic operators, we give assumptions on which our work is based: The input
images used in this work are rectified to have a horizontal epipolar line. 
Consequently, the disparity depends only on the column
index of the pixel: the pixel \emph{$(r,c)$} in the reference image is
paired to the pixel $(r,c+d)$ in the target image. A $3D$
disparity space is defined. Dimension of the disparity space are $r$, $c$ and $d$ 
to designate respectively row, column and disparity. Each element $(r,c,d)$ of
the disparity space is projected to the pixel $(r,c)$ in the reference
image and $(r,c+d)$ in the target image. The element $(r,c,d)$
refers to the pairing of the pixel $(r,c)$ of the reference image
and the pixel $(r,c+d)$ of the target image. Let $R\times C$ be
the image size.

\subsection{New encoding scheme}

In genetic algorithms, a population of individuals representing potential solutions is preserved within the space of search. Each individuals is represented by a finite length vector of components, or variables, by means of some symbols, mostly the binary symbols ${0,1}$. 

To continue the genetic analogy these individuals are likened to chromosomes
and the variables are analogous to genes. Thus a chromosome (potential
solution) is composed of several genes (variables) {[}16{]}.

State of-the-art genetic stereo algorithms provide good matching results, but their
major limitation is the computational time. This disadvantage can be explained by the fact that classical genetic approaches use binary encodings for individuals, which leads to some matching ambiguities. Furthermore, a binary encoding requires more storage
 space and more computing time {[}5{]}. To get around these limitations, we propose a new encoding for individuals, more compact than binary encoding and requiring much less space.

In stereo matching problems, the solution is
a disparity map. A chromosome is an encoding scheme of a potential
solution. Consequently, we define a chromosome as a matrix $C$ with
the same size $(R\times C)$ as the input image. The value of the
cell $(r,c)$ of the chromosome matrix represents the disparity $d$
between the pixel $(r,c)$ in the reference image (the right image)
and the corresponding pixel in the target image.

\begin{figure}
\centering{}\includegraphics[scale=0.85]{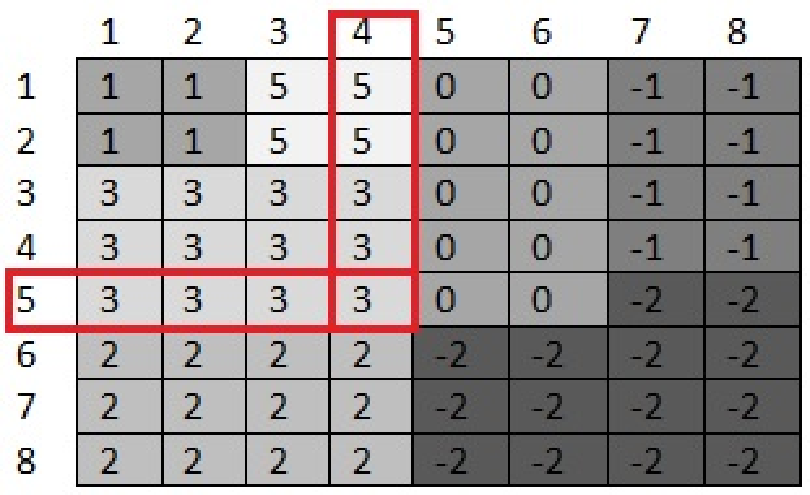}\caption{A chromosome representing a 8x8 image, the value of the element (5,4) is 3 which means that pixel (5,4) in the reference image corresponds to the pixel (5,7) in the target image.}
\end{figure}

\subsection{Fuzzy fitness}

In stereo matching, corresponding pixels are identified using several constraints such as similarity constraint and disparity smoothness. Similarity assumption assumes that the projections of the same punctual area of the scene have nearly the same intensities.

Disparity smoothness constraint assumes that the disparities are smooth in a local neighbourhood. Accordingly, the evaluation function is formulated based on similarity and  smoothness constraints.

Similarity assumption is commonly used by stereo vision approaches to define a static fitness function by calculating the difference of intensities between two neighbourhood of the candidate pixels. A static fitness is not robust to noise caused by changes of illumination conditions, sampling, scanning... In this work, we propose to calculate a fuzzy quantification of the similarity assumption which can model more efficiently non ergodic phenomena. For this purpose, a grey scale classification of pixels is defined {[}14{]}. This classification is based on three classes;  \textit{black pixels}, \textit{white pixels} and \textit{average pixels}. We define membership functions (Eq.1) of theses classes as Gaussian centered in 0, 127.5 and 255.

\begin{equation}
\mu_{class}(m)=exp\left(-\frac{(I(m)-c_{class})^{2}}{2\sigma_{class}^{2}}\right)
\end{equation}

$I(m)$ is the intensity at the pixel $m$, $c_{class}$ and $\sigma_{class}$
are respectively the center and the standard deviation of the class
under consideration. Based on this classification and using similarity
assumption, we can assume that the matching of two pixels $m_{1}$
and $m_{2}$ projections of the same punctual area M on the stereo
images, is "possible" if the two pixels are in the same grey
class, that means ($m_{1}$ is \textbf{black} AND $m_{2}$ is \textbf{black})
OR ($m_{1}$ is \textbf{white} AND $m_{2}$ is \textbf{white}) OR
( $m_{1}$ is \textbf{average} AND $m_{2}$ is \textbf{average}).
Using classical fuzzy operators, we define a fuzzy
\emph{matching} \textit{possibility} metric: $\Pi(m_{1},m_{2})$ (given
by Eq. 2). $\Pi(m_{1},m_{2})$ is a measure of co-membership to a
same grey class. It reflects how much it is "possible" to have
$m_{1}$ and $m_{2}$ as corresponding pixels.

\begin{equation}
\Pi(m_{1},m_{2})=\max\left(\begin{array}{c}
\min(\mu_{black}(m_{1}),\mu_{black}(m_{2})),\\
\min(\mu_{average}(m_{1}),\mu_{average}(m_{2})),\\
\min(\mu_{white}(m_{1}),\mu_{white}(m_{2}))
\end{array}\right)
\end{equation}

$\mu_{class}(m)$ is the degree of membership of the pixel $m$ to
the class under consideration. The possibility of matching ranges
between $0$ and $1$. Thereafter, we will use the notation $\Pi(r,c,d)$=$\Pi(m_{1},m_{2})$
with $m_{1}=(r,c)$ and $m_{2}=(r,c,d)$.

During the evolutionary process, the fittest individual is determined by a fitness function, also known as the evaluation function. This function aims to direct the selection by assigning a fitness score to each chromosome of each generation.

As classical approaches, our fitness function uses intensity
similarity and disparity smoothness. But instead of using the difference
of intensity measurements, which can be easily affected by noise,
we use the matching possibilities. That makes the proposed fuzzy fitness
more robust to noise, change of view point, occlusions... The Fitness
of an individual matrix C is given by Eq. 3.

\begin{equation}
\begin{array}{c}
F(C)=\begin{array}{c}
\\
\sum\\
r,c
\end{array}S_{r,c}\begin{array}{c}
\\
\sum\\
(i,j)\in N
\end{array}\Pi(r+i,c+j,C(r+i,c+j))\\
\\
\end{array}
\end{equation}

\begin{equation}
\begin{array}{c}
S_{r,c}=\left|\nabla(r,c)\right|\left|\nabla(r,c+C(r,c))\right|\end{array}
\end{equation}

$C(r,c)$ is the disparity value of the cell $(r,c)$ within the chromosome
matrix $C$. $N$ is a neighbouring introduced to have a discriminating
comparison between the projections. $\left|\nabla(r,c)\right|$ and
$\left|\nabla(r,c+C(r,c))\right|$ are Sobel gradient norms respectively
on reference pixel $(r,c)$ and target pixel $(r,c+C(r,c))$. That
is intended to penalize pixels which project onto uniform regions,
i.e. less significant pixels.

\subsection{Genetic operators}

Some genetic operators are intended to preserve diversity and to avoid local optima such as mutation. Other operators are used to combine best individuals into others such as crossover {[}7{]}. Offspring is generated from two chosen individuals from the current population by exchanging a part of the matrices (Figure 2). The child chromosome acquires some characteristics from each parent. The crossover
line is chosen randomly in the interval $[1,R]$ for a $R\times C$
image.

\begin{figure}
\centering{}\includegraphics[scale=0.35]{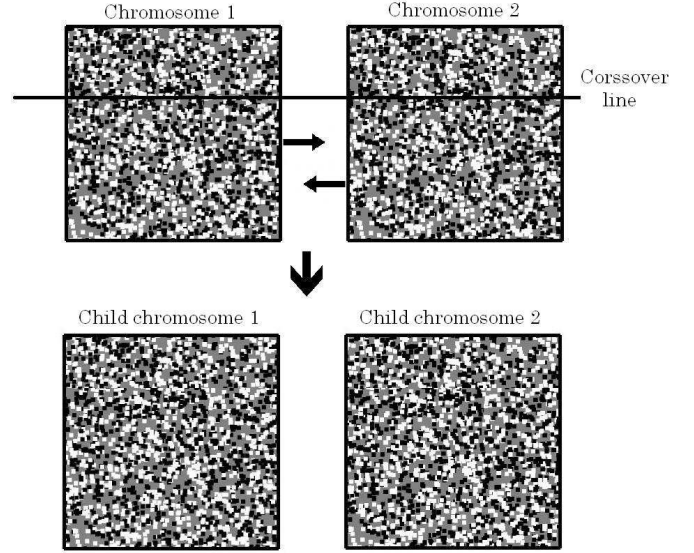}\caption{Example of a crossover of two chromosomes in a random dot matrices}
\end{figure}

Mutation is randomly applied to individuals to produce a varied offspring.
 This operator randomly changes one or several cells (a neighbouring N of uniform cells) in an individual
(Figure 3). Offspring may, thus, inherit several characteristics from
their parents. Mutation prevents falling into local optima
and increases the chance to find global optima. The mutation
rate is set to 40\% and affects matrix cells with lower matching possibilities.

\begin{figure}
\centering{}\includegraphics[scale=0.35]{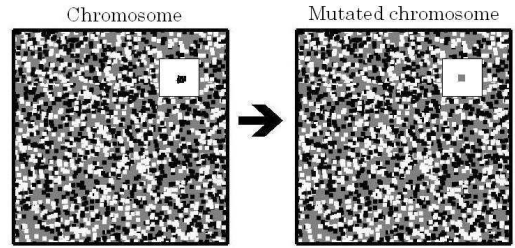}\caption{Example of a chromosome mutation in a random dot matrices}
\end{figure}

\subsection{Selection}

Once the choice of the chromosome encoding is fixed, genetic algorithm starts by generating the initial population. Subsequently, a varied offspring is produced using crossover and mutation operators. All the individuals in a given generation are evaluated by the proposed fuzzy fitness function. Based on this evaluation, some chromosomes are selected to play the role of parents in the next generation. This iterative process is repeated until a stop condition is reached. 

In our algorithm, we use an elitist deterministic selection. Chromosomes are ordered according to their evaluation score and only the fittest individuals are retained (around 40\%). The fitness used to evaluate all individuals is given by Eq. (3). The fitness of an individual represents his survival probability to be selected for the next generation. Sigmoid or rank functions are commonly used to convert from fitness to a survival probability. Selection is controlled by a random function which generates values between 0 and 1 for each selection. If the fitness of the current individuals is greater than the generated probability, then this individual will be chosen for the next generation. The evolution process is ended when the fittest individual of the current generation satisfies the stop condition.

\section{Experimental results}

In this work,  we acquire images using a pre-calibrated stereo rig. We also consider that the reference and
target images are rectified so that epipolar lines are horizontal
and have the same line number in the reference and target images.
 We assume that the rectification is correct,
which means that the disparity depends only on the column index of
the pixel: the pixel $(r,c)$ in the reference image is matched to the
pixel $(r,c+d)$ in the target image.

The performance of our approach was assessed on the Middlebury benchmark
site {[}12{]}, which is well recognized and widely used by the computer
vision community. The site contains standard datasets for researchers
to experiment with using an on-line interface. Also, it contains results
for the most current algorithms and their relative ranking.

First pair of stereo images in Figure 4 shows the Teddy stereo pair
and the ground truth in (A), the disparity maps obtained with different
parameters in (B).

\begin{figure}
\begin{centering}
\includegraphics[scale=0.5]{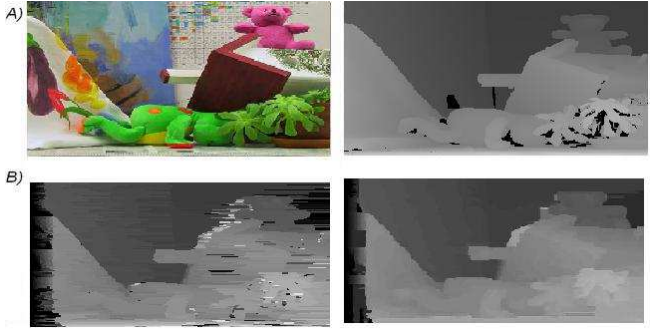}
\par\end{centering}

\caption{A) Reinforce image of Teddy (left) and Ground truth map (right), B)
Disparity map after 10 generations with initial population of 40 individuals
(Left), Disparity map after 100 generations with initial population
of 70 individuals (Right).}
\end{figure}

Figure 5 below presents some results in test images. The top two images
(A) are the reference images of Sawtooth and Venus, the two images
in (B) are ground truth (of the disparity map) and the bottom two
images correspond to the disparity maps obtained (on the Middlebury
site) by our approach after 80 generations. All of the performance
evaluations were conducted on-line on the Middlebury evaluation site
{[}12{]}, according to the benchmark protocol and the test datasets
(of 4 stereo pairs) agreed upon. The error rates obtained in this
evaluation were 15.3\%. Note that all disparity maps were obtained
with an initial population of 70 individuals and a mutation rate of
40\%, used in all of the accuracy evaluation test data on the Middlebury
site.

\begin{figure}[H]
\begin{centering}
\includegraphics[scale=0.5]{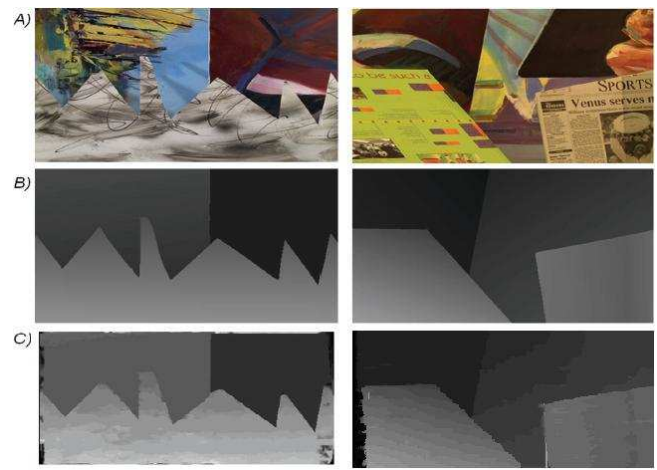}
\par\end{centering}

\caption{A) Reference images of the Sawtooth (left) and Venus (right) stereo
pair, B) Ground truth disparity maps, C) Disparity maps after 100
generations}
\end{figure}

A main evaluation criterion used is the number of incorrect pixels,
i.e., pixels for which the absolute disparity difference between the
solutions provided and the ground truth is greater than 1. Table I
shows the percentage of incorrect pixels obtained via the Middlebury
evaluation site for our approach.

\begin{table}
\begin{centering}
\begin{tabular}{ccccc}
 & Tsukuba & Venus & Teddy & Cones\tabularnewline
\hline 
error rate & 2,74\% & 2,5\% & 14,7\% & 7,78\%\tabularnewline
\hline 
\end{tabular}
\par\end{centering}

\caption{Error rate, according to the Middlebury evaluation site for our approach}
\end{table}

To check the efficiency of our algorithm, we performed a number of tests using the 
\textit{sawtooth} image. Three genetic stereo matching algorithms (\textit{\emph{Table
}}II) are considered for the comparison. The experiment is made on a dual
core with 3GHz and 1Go of RAM. Table II illustrates the computational time for
\textit{sawtooth} with a half-size image (217x190). We have
used an initial population of 50 individuals and the maximum number
of generations is set to 10 in this test. Table II shows
the improvement in computational time of our algorithm compared
to state of-the-art algorithms. Results show that our algorithm is suitable for real-time application.
\begin{table}
\begin{centering}
\begin{tabular}{cc}
Algorithms & Execution time\tabularnewline
\hline 
\hline 
\textit{\emph{Our algorithm}} & 3.4 s\tabularnewline
\textit{\emph{Han's algorithm {[}11{]}}} & 9.7 s\tabularnewline
Dong's algorithm {[}10{]} & 5 s\tabularnewline
Nguyen's algorithm {[}13{]} & 2.7 s\tabularnewline
\hline 
\end{tabular}
\par\end{centering}

\caption{Comparison of execution time in seconds with existing stereo matching
algorithms for \textit{sawtooth} image with a half-size image (217x190)
after 10 generations.}
\end{table}

\section{Conclusion}

We presented in this paper a genetic fuzzy algorithm for the stereo matching
problem. In this algorithm a new encoding scheme was proposed. A fuzzy
formulation of the similarity assumption is used to propose a matching
possibility metric. A fuzzy fitness function using matching possibilities
and disparity smoothness is defined to evaluate individuals. According
to the proposed encoding scheme, genetic operators were adapted. Experimental
results show that the presented algorithm yields a significant improvement
in accuracy and computational time relative to state of-the-art genetic
stereo matching algorithm.

\end{document}